\documentclass{article}
\usepackage{spconf}
\usepackage{hyperref}       
\usepackage{url}            
\usepackage{booktabs}       
\usepackage{amsfonts}       
\usepackage{nicefrac}       
\usepackage{microtype}      
\usepackage{xcolor}         
\usepackage{latexsym}
\usepackage{graphicx}
\usepackage{tikz}
\usepackage{amsthm,algorithmic}
\usepackage{epsfig,psfrag,amstext,amsfonts,amssymb,subeqnarray,color}
\usepackage{epstopdf,bm,multirow}
\usepackage{amsmath}
\usepackage{array,makecell,bm,threeparttable}  
\usepackage{marvosym}
\usepackage{ifsym}
\usepackage{textcomp}
\usepackage{hyperref}
\usepackage{adjustbox}
\usepackage{nccfoots}
\usepackage[linesnumbered,ruled]{algorithm2e}
\usepackage{booktabs}
\usepackage{cuted}
\usepackage{cite}
\usepackage{pifont}
\usepackage{flushend} 
\usepackage[caption=false,font=footnotesize,subrefformat=parens,labelformat=parens]{subfig}

\newtheorem{remark}{Remark}

\def\x{{\mathbf x}}
\def\y{{\mathbf y}}
\def\z{{\mathbf z}}
\def\f{{\mathbf{f}}}

\def\u{{\mathbf u}}
\def\m{{\mathbf m}}

\def\vx{{\vec{\mathbf x}}}
\def\vy{{\vec{\mathbf y}}}
\def\vz{{\vec{\mathbf z}}}
\def\vf{{\vec{\mathbf f}}}
\def\vu{{\vec{\mathbf u}}}

\def\k{{\boldsymbol{K}}}

\def\J{{\mathbf J}}
\def\G{{\mathbb G}}

\def\cX{{\cal X}}

\def\cD{{\cal D}}

\def\cN{{\cal N}}

\def\bzeta{{\bm{\zeta}}}
\def\btheta{{{\bm{\theta}}}}

\definecolor{orange}{RGB}{200,0,100}

\newcommand{\cred}[1]{{\color{red}#1}}

\title{
Towards Efficient Modeling and Inference in Multi-Dimensional \\ Gaussian Process State-Space Models
\vspace{-.05in}
} 
\name{
Zhidi Lin\textsuperscript{$\ast$ } \quad
Juan Maro\~{n}as\textsuperscript{$\dagger$} \quad
Ying Li\textsuperscript{$\ddagger$} \quad 
Feng Yin\textsuperscript{$\ast$}(\textrm{\Letter}) \quad
Sergios Theodoridis\textsuperscript{$\sharp$}
\thanks{
This work was supported by the NSFC with Grant No. 62271433 and in part by Guangdong Research Project with Grant No. 2017ZT07X152.
The corresponding author is Feng Yin (\textit{yinfeng@cuhk.edu.cn}). 
 }
\vspace{-.05in}
}
\address{
$\ast$
The Chinese University of Hong Kong, Shenzhen, China 
\ \ $\ddagger$ The University of Hong Kong, HKSAR 
\\
$\dagger$ Machine Learning Group, Autonomous University of Madrid, and Cognizant, Madrid, Spain
\\
$\sharp$ National and Kapodistrian University of Athens, Greece, and Aalborg University, Denmark
\vspace{-.1in}
}
\begin{document}
\topmargin=0mm
\ninept
\maketitle
\begin{abstract}  \vspace{-.07in}
The Gaussian process state-space model (GPSSM) has attracted extensive attention for modeling complex nonlinear dynamical systems. However, the existing GPSSM employs separate Gaussian processes (GPs) for each latent state dimension, leading to escalating computational complexity and parameter proliferation, thus posing challenges for modeling dynamical systems with high-dimensional latent states. To surmount this obstacle, we propose to integrate the efficient transformed Gaussian process (ETGP) into the GPSSM, which involves pushing a shared GP through multiple normalizing flows to efficiently model the transition function in high-dimensional latent state space. Additionally, we develop a corresponding variational inference algorithm that surpasses existing methods in terms of parameter count and computational complexity. Experimental results on diverse synthetic and real-world datasets corroborate the efficiency of the proposed method, while also demonstrating its ability to achieve similar inference performance compared to existing methods. Code is available at \url{https://github.com/zhidilin/gpssmProj}.
\end{abstract}
\vspace{-.05in}
\vspace{-.1in}
\section{Introduction}
\label{sec:intro} \vspace{-.1in}

Gaussian process state-space models (GPSSMs) \cite{frigola2013bayesian} have gained significant popularity among data-driven state-space models (SSMs) \cite{krishnan2017structured,alaa2019attentive,frigola2013bayesian,frigola2014variational,frigola2015bayesian,doerr2018probabilistic,ialongo2019overcoming}, due to their capability of integrating non-parametric Bayesian Gaussian processes (GPs) \cite{williams2006gaussian,suwandi2022gaussian,yin2020linear} as function priors within the classical SSM \cite{sarkka2013bayesian}. This integration empowers the model to effectively learn the system dynamcis from noisy measurements with explicit uncertainty calibrations \cite{cheng2022rethinking}. Additionally, GP is able to automatically scale model complexity based on data volume \cite{theodoridis2020machine}. Consequently, GPSSMs and their variants have demonstrated successful applications in diverse domains, including human motion capture, pedestrian tracking, and navigation \cite{wang2007gaussian,xie2020learning,yin2020fedloc,zhao2019cramer}. Research efforts have also focused on advancing the simultaneous learning and inference capabilities of GPSSMs \cite{frigola2013bayesian, frigola2014variational,frigola2015bayesian,eleftheriadis2017identification,doerr2018probabilistic,ialongo2019overcoming,lindinger2022laplace,lin2022output, lin2023towards,liu2021gaussian,fan2023free}.

However, in the context of high-dimensional latent state spaces, all the existing GPSSMs face two primary challenges. First, current GPSSM methods often resort to independent GPs for modeling multiple outputs of transition functions, aiming for simplicity but overlooking their dependencies. This disregard for dependencies can lead to a model mismatch and the loss of inductive bias among the outputs \cite{chen2022multitask}. This can ultimately impede the model's generalization capacity and cause a decline in inference performance, particularly when latent states are only partially observed \cite{lin2022output}. Second, employing separate GPs to model the state transition for each latent state dimension leads to a quadratic expansion in the number of parameters, coupled with a linear rise in $\mathcal{O}(n^3)$ computational complexity, as dimensionality increases, see Fig.~\ref{fig:EGPSSM_results}. Here, $n$ represents the sample count used for computing the GP kernel matrix \cite{williams2006gaussian}. This escalating computational burden and parameters proliferation can become prohibitively cumbersome, especially when dealing with high-dimensional latent spaces. Consequently, addressing these two challenges becomes imperative to enhance the applicability and scalability of GPSSM in practical applications. 
For the first challenge, existing approaches have explored potential solutions, with some using a linear model of coregionalization (LMC)-based multi-output GP to model this correlation \cite{lin2022output}. The transformed GP (TGP) \cite{maronas2021transforming} framework has also been introduced, wherein multiple independent GPs are transformed by a normalizing flow \cite{kobyzev2020normalizing} to somewhat obtain correlated outputs \cite{lin2023towards}. Despite these efforts, the persistent challenge of escalating complexity remains an obstacle across existing works, necessitating research to further enhance the applicability of GPSSMs in high-dimensional latent state spaces.

This paper aims to address the escalating computational complexity and parameter proliferation in the GPSSM while introducing a novel form of output dependence. The main contributions are summarized as follows.
First, we present an innovative efficient GPSSM paradigm that deviates from the standard approach of employing separate GPs for each one of the latent dimensions. Instead, we adopt the efficient transformed GP (ETGP) \cite{maronas2023efficient}, capitalizing on multiple normalizing flows \cite{kobyzev2020normalizing} to enact transformations on a shared GP across each dimension of the latent state space. This strategic shift allows us to attain streamlined modeling while effectively establishing output dependencies. 
Second, for joint learning and inference in the proposed efficient GPSSM,  we propose a proficient sparse GP \cite{hensman2013gaussian}-based variational algorithm that enhances computational efficacy and streamlines parameter scale. 
Third, experimental results, obtained using real and synthetic datasets, corroborate comparable performance of the proposed efficient GPSSM to existing GPSSMs, albeit at substantial reductions in both computational complexity and parameter count.

The remainder of this paper is organized as follows. Some preliminaries related to GPSSM are provided in Section \ref{sec:preliminaries}.  Section \ref{sec:proposed-model} introduces our proposed efficient output-dependent GPSSM and the associated learning and inference algorithm. Numerical results are provided in Section \ref{sec:experimental-results}. Finally, we conclude the paper in Section \ref{sec:conclusion}.
\vspace{-.1in}

\section{Preliminaries}
\label{sec:preliminaries} 

\vspace{-.08in}
\subsection{Gaussian Processes (GPs)}
\label{subsec:GP-review}  
\vspace{-.1in}
A GP defines a collection of random variables indexed by $\cX \subseteq \mathbb{R}^{d_x}$, where any finite subset of these variables follows a joint Gaussian distribution \cite{williams2006gaussian}. Typically, a GP is employed to represent a distribution over random functions $f(\x): \mathbb{R}^{d_x} \mapsto \mathbb{R}$, given by:
\begin{equation}
\setlength{\abovedisplayskip}{3.5pt}
\setlength{\belowdisplayskip}{3.5pt}
f(\x) \sim \mathcal{GP}\left(\mu(\x), \ k_{\boldsymbol{\theta}_{gp}}\left(\x, \x^{\prime} \right)\right),
\end{equation}
where $\mu(\x)$ is a mean function, often set to zero in practice; $k_{\boldsymbol{\theta}_{gp}}\left(\x, \x^{\prime}\right)$ is a covariance/kernel function; $\boldsymbol{\theta}_{gp}$ represents a set of hyperparameters that are tuned for model selection. In the rest of the paper, $\btheta_{gp}$ is omitted for the sake of notation brevity. By applying Bayes' theorem, the function prior is combined with new data to obtain an analytical posterior distribution. Specifically, given a noise-free training dataset $\cD \!=\! \{X, \f\}\!=\!\{\x_{i}, \mathrm{f}_i\}_{i=1}^n$, the posterior distribution $p(f(\x_*) \vert \x_*, \cD)$ at any test input $\x_* \!\in\! \cX$ follows a Gaussian distribution, fully characterized by the posterior mean $\xi$ and the posterior variance $\Xi$:
\begin{subequations}
\setlength{\abovedisplayskip}{4pt}
\setlength{\belowdisplayskip}{4pt}
\begin{align}
& \xi(\x_*) = \bm{K}_{{\x_*}, X} \bm{K}_{X,X}^{-1} \f, \label{eq:posterior_mean} \\
& \Xi(\x_*) = k(\x_*, \x_*) - \bm{K}_{{\x_*}, X} \bm{K}_{X,X}^{-1} \bm{K}_{{\x}_*, X}^\top, \label{eq:posterior_variance}
\end{align}
\end{subequations}
where $\bm{K}_{X,X}$ represents the covariance matrix evaluated on the training input ${X}$, with each entry given by $[\bm{K}_{X,X}]_{i,j} = k({\x}_i, {\x}_j)$; $\bm{K}_{{\x}*, X}$ denotes the cross-covariance matrix between the test input ${\x}_*$ and the training input ${X}$.

\vspace{-.13in}
\subsection{Gaussian Process State-Space Model (GPSSM)}
\label{subsec:GPSSM-review}  \vspace{-.07in}
A generic state-space model (SSM) characterizes the probabilistic relationship between the latent state, $\x_t \in \mathbb{R}^{d_x}$, and the observation, $\y_t \in \mathbb{R}^{d_y}$. Mathematically, it is represented as follows:
\begin{equation}
\label{eq:SSM}
\setlength{\abovedisplayskip}{4pt}
\setlength{\belowdisplayskip}{4pt}
\x_{t+1} = f(\x_t) + \mathbf{v}_t,  \qquad  \y_{t} = g(\x_t) + \mathbf{e}_t,
\end{equation} 
where $\mathbf{v}_t$ and $\mathbf{e}_t$ are additive noise terms, and $f(\cdot)$ and $g(\cdot)$ are referred to as the \textit{transition function} and \textit{emission function}, respectively. 

Incorporating a GP prior over the transition function $f(\cdot)$ and assuming a parametric emission function $g(\cdot)$ in the classic SSM (see Eq.~\eqref{eq:SSM}) leads to the well-known GPSSM\footnote{The GPSSM considered in this paper keeps the same model capacity as the \textit{ones with both transition and emission GPs} while avoiding the severe \textit{non-identifiability} issue.  One can refer to \cite{frigola2015bayesian} (Section 3.2.1) for more details.}\!  \cite{frigola2015bayesian}, expressed as follows:\!
\begin{subequations}
\setlength{\abovedisplayskip}{4pt}
\setlength{\belowdisplayskip}{4pt}
    \label{eq:gpssm}
    \begin{align}
        &  f(\cdot)  \!\sim\! \mathcal{G} \mathcal{P}\left(\bm{0}, \bm{k}(\cdot, \cdot)\right), \quad {\f}_{t} \!=\!f(\mathbf{x}_{t-1}), \quad \mathbf{x}_{0} \!\sim\! p(\mathbf{x}_{0}),  \\
        &  \quad \mathbf{x}_{t} \vert {\f}_{t}  \!\sim\! \mathcal{N}\left( \x_t \vert {\f}_{t}, \mathbf{Q}\right), \quad  \mathbf{y}_{t} \vert  \mathbf{x}_{t} \!\sim\! \cN \left(\y_t \vert \bm{C} \mathbf{x}_{t}, \mathbf{R}\right),
    \end{align}
\end{subequations}
where the emission model is assumed to be known and linear and the coefficient matrix, $\bm{C} \!\in\! \mathbb{R}^{d_y \times d_x}$, mitigates the system non-identifiability \cite{frigola2015bayesian}. The initial state prior distribution $p(\x_0)$ is also known and assumed to follow a Gaussian distribution. Both state transitions and observations are corrupted by zero-mean Gaussian noise with covariance matrices $\mathbf{Q}$ and $\mathbf{R}$, respectively.  In the case of state dimension $d_x\!>\!1$, the transition $f(\cdot): \mathbb{R}^{d_x} \mapsto  \mathbb{R}^{d_x}$ is typically modeled with $d_x$ mutually independent GPs. Each independent GP represents a dimension-specific function $f_d(\cdot): \mathbb{R}^{d_x} \mapsto \mathbb{R}$, and the multivariate output is denoted as 
\begin{equation}
\setlength{\abovedisplayskip}{4pt}
\setlength{\belowdisplayskip}{4pt}
\f_t = f(\x_{t-1}) \triangleq \{f_d(\x_{t-1})\}_{d=1}^{d_x} \triangleq \{\f_{t,d}\}_{d=1}^{d_x} , 
\label{eq:multivariate_GP}
\end{equation}
where each independent GP is associated with a unique kernel function and the associated hyperparameters. The challenging task of GPSSM lies in simultaneously learning the transition function and the noise parameters, i.e., learning $[\btheta_{gp}, \mathbf{Q}, \mathbf{R}]$, while inferring the latent states of interest.  

Despite the popularity of the aforementioned GPSSM modeling approach, it is plagued by two significant caveats as outlined in Section \ref{sec:intro}, namely the escalating burden of complexity (encompassing computational cost and size of the parameter space), and the model output independence. In the following section, we will present our novel solution that tackles the challenge of growing complexity while also establishing output dependence.

\vspace{-.1in}
\section{Efficient GPSSM Modeling and Inference}
\label{sec:proposed-model}
\vspace{-.1in}
Section \ref{subsec:ETGPSSMs} introduces our novel efficient GPSSM formulation tailored for multi-dimensional latent state spaces. Section \ref{subsec:Inference_ETGPSSM} presents the proposed variational algorithm for learning and inference in the efficient GPSSM.
\vspace{-.05in}
\subsection{Efficient GPSSM Modeling}
\label{subsec:ETGPSSMs} \vspace{-.02in}
In the context of a multi-dimensional latent state space, our GPSSM formulation departs from the conventional use of $d_x$ separate GPs for modeling the transition function. Instead, we incorporate a more efficient approach, ETGP \cite{maronas2023efficient}. This entails utilizing a single GP and employing normalizing flow \cite{kobyzev2020normalizing} to transform the process within each dimension of the latent state. It is noteworthy that the ETGP has already showcased successful applications in large-scale multi-classification scenarios \cite{maronas2023efficient}.   In essence, the new GPSSM formulation can be expressed as:
\begin{subequations}
\setlength{\abovedisplayskip}{4pt}
\setlength{\belowdisplayskip}{4pt}
\label{eq:ETGPSSM}
\begin{align}
    &  \quad \Tilde{f} \!\sim\! \mathcal{GP}(0, k(\cdot, \cdot),  \ \ \Tilde{\mathrm{f}}_t \!=\! \Tilde{f}(\x_{t-1}),  \ \  {\f}_t \!=\! \{ \G_{\btheta_d}(\Tilde{\mathrm{f}}_t) \}_{d=1}^{d_x}, \label{subeq:ETGPSSM_a}\\
    & \mathbf{x}_0 \!\sim\! p(\x_0), \ \  \mathbf{x}_{t} \vert \f_{t}  \!\sim\! \mathcal{N}(\x_t \vert \f_{t}, \mathbf{Q}), \ \ \mathbf{y}_{t} \vert  \mathbf{x}_{t} \!\sim\! \cN ( \y_t \vert \bm{C} \mathbf{x}_{t}, \mathbf{R}),
\end{align}
\end{subequations}
where the dimension-specific normalizing flow, $\G_{\btheta_{d}}(\cdot)$, comprises a set of invertible and differentiable transformations, $  \{ \G_{\theta_{d,j}}(\cdot): \mathbb{R} \mapsto \mathbb{R} \}_{j=0}^{J-1}$, parameterized by $\bm{\theta}_{d} \triangleq \{\theta_{d,j-1} \}_{j=1}^J, J \in \mathbb{N}$. Due to these time-wise mappings, any finite $T$-dimensional multivariate random variable $\vf \! \triangleq \! \f_{1:T} \! \triangleq \! \{\f_t\}_{t=1}^T$ is guaranteed from the induced Gaussian copula processes \cite{wilson2010copula}.
\begin{remark}
The new GPSSM formulation presented in Eq.~\eqref{eq:ETGPSSM} maintains constant computational complexity when the computational cost of the normalizing flows becomes negligible compared to the GP part. Consequently, the overall computational burden aligns with that of calculating a single GP, resulting in a highly efficient method. 
Another advantage is the establishment of dependency among the resulting $d_x$ Gaussian copula processes \cite{wilson2010copula}. This dependency arises due to their shared basis GP, denoted as $\Tilde{f}$, while simultaneously exhibiting distinct marginal distributions \cite{maronas2023efficient}. Such correlated yet individually varying processes enhance the modeling capacity and enable a more comprehensive representation of complex relationships within the latent state space.
\end{remark}
In principle, the GPSSM can integrate diverse forms of normalizing flows, spanning from basic and interpretable elementary flows to advanced flows developed in recent years \cite{kobyzev2020normalizing}. This paper investigates two commonly used elementary flows. The first one is the Sinh-Arcsinh-Linear (SAL) flow \cite{maronas2021transforming,rios2019compositionally}, which can be applied by stacking $J\in \mathbb{N}$ layers, with each layer represented as follows:
\begin{equation}
    \label{eq:SAL_flows}
    \G_{\theta_{d,j}}(\cdot) = \alpha_{d,j} \sinh \left(\varphi_{d,j} \operatorname{arcsinh}(\cdot ) - \gamma_{d,j} \right)+\beta_{d,j}, 
\end{equation}
where $\theta_{d, j} \!\triangleq\! [\alpha_{d,j}, \beta_{d,j}, \gamma_{d,j}, \varphi_{d,j}]$, $d \!\in \! \{1, 2, \ldots, d_x\}$, $j \!\in \! \{0, 1, \ldots, J\!-\!1\}$.  The SAL flow can be employed to control the mean, variance, asymmetry, and kurtosis of the function priors \cite{rios2019compositionally}. Another one is the simple linear flow,  that is, 
\begin{equation}
 \f_{t, d} = \alpha_d \cdot \Tilde{\mathrm{f}}_t + \beta_d, \ d \!\in \! \{1, 2, \ldots, d_x\},
\end{equation}
with $\btheta_d = [\alpha_d, \beta_d]$. When using linear flow, it is straightforward to see that the $d_x$ outputs are dependent GPs (see Proposition 1 in \cite{maronas2023efficient}), such that for any two dimensions $d$ and  $d^\prime$, we have
\begin{equation}
\begin{aligned}
    & \mathbb{E}[\f_{t,d}] = \beta_d,  \ \ 
    \text{Cov}[\f_{t,d}, \f_{t^\prime,d^\prime} ]  = \alpha_d \alpha_{d^\prime} k(\x_{t-1}, \x_{t^\prime-1}).
\end{aligned}
\end{equation}
Both of these commonly used normalizing flows exhibit linear runtime complexity and involve only a small amount of parameters, whose computational effect is practically negligible when contrasted with the complexity of separate GPs in existing GPSSMs. More importantly, the computation of the $d_x$ flows can be parallelized through batched matrix operations, in cases where the flows share the same functional form, which leads to a substantial enhancement in model efficiency.

\subsection{Efficient GPSSM Inference}
\label{subsec:Inference_ETGPSSM}  \vspace{-.1in}
In addition to the computational efficiency enhancement in the efficient GPSSM, as outlined in Section \ref{subsec:ETGPSSMs}, we employ sparse GP methods \cite{hensman2013gaussian} to further alleviate the computational load that stems from the GP model with a large sample count. This involves introducing a compact set of inducing points $\vec{\z} \triangleq \{\z_{i}\}_{i=1}^m$ and $\vec{\u} \triangleq \{\u_{i}\}_{i=1}^{m}, m\ll T,$ that act as a surrogate for the corresponding GP. Here, $\u_{i} = \Tilde{f}(\z_{i}) \in \mathbb{R}$ and $\z_i \in \mathbb{R}^{d_x}$. To facilitate discussions, we define $\vx \triangleq \{\x_{t}\}_{t=0}^{T}$ and $\vy \triangleq \{\y_{t}\}_{t=1}^{T}$.  Based on these configurations, the joint distribution of the proposed GPSSM, augmented with inducing points, is
\begin{equation}
\setlength{\abovedisplayskip}{3.5pt}
\setlength{\belowdisplayskip}{3.5pt}
\begin{aligned}
\!\! p(\vy, \vx, \vf, \vu) \!=\! p(\x_{0}) \prod_{t =1}^{T}  p(\y_{t} \vert \x_{t}) p(\x_{t} \vert \f_{ t})p(\f_{t} \vert \x_{t-1}, \vu) p(\vu),
\end{aligned}
\end{equation}
where $p(\vu) = \cN(\vu \vert \mathbf{0}, \bm{K}_{\vz, \vz})$, and the distribution of the transition function outputs $p(\f_{t} \vert \x_{t-1}, \vu)$ are determined by the shared GP $p(\Tilde{\mathrm{f}}_t \vert \x_{t-1}, \vu)$ and the normalizing flows, where
\begin{equation}
\setlength{\abovedisplayskip}{4pt}
\setlength{\belowdisplayskip}{4pt}
p(\Tilde{\mathrm{f}}_t ~\vert~ \x_{t-1}, \vu) \!=\! \cN\left(\Tilde{\mathrm{f}}_t \mid {\xi}(\x_{t-1}),\ {\Xi}(\x_{t-1}) \right)
\end{equation}
is the GP posterior distribution with $\x_{t-1}$ being the test input while $(\vz, \vu)$ being the training data, see Eqs.~(\ref{eq:posterior_mean}) and (\ref{eq:posterior_variance});
Therefore, 
\begin{equation}
\setlength{\abovedisplayskip}{4pt}
\setlength{\belowdisplayskip}{4pt}
    p(\f_{t} \vert \x_{t-1}, \vu) \!=\!  \underbrace{p(\mathrm{\Tilde{f}}_t \vert \x_{t-1}, \vu) \cdot \J_\f}_{=p(\f_{t,1} \vert \vu, \x_{t-1})} \cdot \prod_{d=2}^{d_x} \underbrace{\delta\left( \f_{t, d} - \G_{\btheta, d} \circ \G_{\btheta, 1}^{-1}(\f_{t, 1}) \right)}_{= p(\f_{t, d} \vert \f_{t, 1}, \vu, \x_{t-1})},
    \label{eq:transition_prior}
\end{equation}
where
$
	 \J_{\f} \triangleq  \prod_{j=1}^{J-1}\left|\operatorname{det} \frac{\partial \ \mathbb{G}_{\theta_{1, j}}\left( {\G_{\theta_{1, j-1}}\left( \cdots \G_{\theta_{1,0}}(\tilde{\mathrm{f}}_{t}) \cdots \right)} \right)}{\partial \  {\G_{\theta_{1, j-1}}\left( \cdots \G_{\theta_{1, 0}}(\tilde{\mathrm{f}}_{t}) \cdots \right)}  }\right|^{-1},
$ and $\delta(\cdot)$ denotes the Dirac measure. Note that this decomposition is not unique. We designate the first dimension, $\f_{t, 1}$, as the \textit{pivot}; and the joint distribution can be expressed equivalently with respect to any other pivot $\f_{t, d}$ for the case $d \!\neq \!1$ (see details in Appendix A, \cite{maronas2023efficient}).

Learning and inference within the GPSSM is plagued by the intractability of $p(\vy) \!=\! \int p(\vy, \vx, \vf, \vu) \mathrm{d}\vx \mathrm{d}\vf \mathrm{d}\vu$. In order to surmount this challenge, instead of relying on Monte Carlo-based methods \cite{frigola2013bayesian}, this paper embraces alternative variational inference techniques \cite{cheng2022rethinking,theodoridis2020machine}.
Variational inference methods involve approximating the intractable posterior distribution $p(\vec{\x}, \vec{\f}, \vec{\u} \vert \vy) \!=\! \frac{p(\vy, \vec{\x}, \vec{\f}, \vec{\u})}{p(\vy)}$ with a variational distribution $q(\vec{\x}, \vec{\f}, \vec{\u})$, leading to a learning and inference objective function, the evidence lower bound (ELBO), denoted as $\mathcal{L}$,
\begin{equation}
\setlength{\abovedisplayskip}{4pt}
\setlength{\belowdisplayskip}{4pt}
    \mathcal{L} \triangleq \mathbb{E}_{q(\vec{\x}, \vec{\f}, \vec{\u})}\left[ \log \frac{p(\vy, \vx, \vf, \vu)}{q(\vec{\x}, \vec{\f}, \vec{\u})}\right] \le \log p(\vy).
\end{equation}
 The choice of the variational distribution influences the tightness of the ELBO and, in turn, determines the learning algorithm for GPSSM \cite{frigola2015bayesian}.
In this paper, we adopt a specific form for the variational distribution $q(\vec{\x}, \vec{\f}, \vec{\u})$ given by
$ q(\x_{0}) \prod_{t =1}^{T} p(\x_{t} \vert \f_{t}) p(\f_{t} \vert \x_{t-1}, \vu) q(\vu), $
where $q(\vu) = \cN(\vu \vert \m, \mathbf{S})$ with $\m \in \mathbb{R}^{m}$ and the covariance matrix $\mathbf{S} \in \mathbb{R}^{m \times m}$ representing the free variational parameters. 
Similarly, the variational distribution for the initial state $\x_0$ is given by $q(\x_{0}) = \mathcal{N}(\x_0 \vert \m_{\x_{0}}, \mathbf{S}_{\x_{0}})$. Here, $\m_{\x_{0}}$ and $\mathbf{S}_{\x_{0}}$ can be treated as free variational parameters or learned through an amortized recognition network \cite{eleftheriadis2017identification} with inputs $\vy$ and parameters $\bzeta$.
Note that the variational distribution form adopted in this paper bears a resemblance to the form used in probabilistic recurrent SSM (PRSSM) \cite{doerr2018probabilistic} and the output-dependent GPSSM (ODGPSSM) \cite{lin2022output}. This deliberate decision facilitates a direct comparison with these methods, enabling us to thoroughly evaluate and analyze the advantages of our GPSSM modeling approach.

With the proposed variational distribution, $q(\vec{\x}, \vec{\f}, \vec{\u})$, and after performing certain algebraic calculations, the ELBO is obtained as,
\begin{equation}
\setlength{\abovedisplayskip}{4pt}
\setlength{\belowdisplayskip}{4pt}
    \begin{aligned}
    \mathcal{L}
    & = \sum_{t=1}^T \mathbb{E}_{q(\x_t)} \left[\log p(\y_t \vert \x_t)\right] - \operatorname{KL}\left(q(\vu) \| p(\vu)\right) \\
    &  \qquad - \operatorname{KL}\left(q(\x_0) \| p(\x_0)\right), 
\end{aligned}
\end{equation}
where the first term encourages the latent states drawn from the variational distribution, 
$$
\setlength{\abovedisplayskip}{4pt}
\setlength{\belowdisplayskip}{4pt}
q(\x_{t})\!=\! \int\! q(\x_0) \prod_{\tau=1}^{t} p(\x_\tau \vert \f_\tau) p(\f_\tau \vert \x_{\tau-1}, \vu) q(\vu) \ \mathrm{d} \x_{0:t-1} \mathrm{d} \vf \mathrm{d} \vu,
$$
to fit the emission model effectively, ensuring a good match between the observed data and the latent states. The second and third terms act as regularizations, controlling the posterior distributions of the initial state and the shared GP (and consequently, the posterior of ETGP $f(\cdot)$). These regularization terms play a crucial role in managing the model complexity and preventing overfitting.

The two regularization terms can be computed analytically due to the Gaussian nature of the distributions involved. However, the expectation terms, $\mathbb{E}_{q(\x_{t})} \left[ \log p(\y_{t} \vert \x_{t})\right], \forall t,$ 
need to be evaluated by the sampling method and reparametrization trick \cite{kingma2019introduction} due to the intractability of $q(\x_{t})$ \cite{doerr2018probabilistic}. 
The samples, $\x_t$, can be obtained as follows.
We first use the reparametrization trick to sample the (1-D) GP function value, $\tilde{\mathrm{f}}_t$, from $q(\tilde{\mathrm{f}}_t \vert \x_{t-1})$ by conditioning on the latent state $\x_{t-1}$, where
\begin{equation}
\setlength{\abovedisplayskip}{4pt}
\setlength{\belowdisplayskip}{4pt}
\begin{aligned}
    q(\tilde{\mathrm{f}}_t \vert \ \x_{t-1}) \! = \! \mathbb{E}_{q(\vu)}  [ p(\tilde{\mathrm{f}}_t \vert \ \x_{t-1}, \vu) ] \! = \!\cN(\tilde{\mathrm{f}}_t \vert \  {\mu}_{t|t-1}, \mathrm{S}_{t|t-1} ),
\end{aligned}
\label{eq:state_conditional}
\end{equation}
with $\mu_{t|t-1} = \k_{\x_{t-1}, \vz} \ \k_{\vz, \vz}^{-1} \ \m,$ and 
$$
\setlength{\abovedisplayskip}{4pt}
\setlength{\belowdisplayskip}{4pt}
\mathrm{S}_{t|t-1}\!=\! k({\x_{t\!-\!1},\x_{t\!-\!1}}) \!-\! \k_{\x_{t\!-\!1}, \vz} \ \k_{\vz, \vz}^{-1} \! \left[\k_{\vz, \vz} \!-\! \mathbf{S} \right]\!\k_{\vz, \vz}^{-1} \ \k_{\x_{t\!-\!1}, \vz}^\top.
$$
We then push the sampled function value $\tilde{\mathrm{f}}_t$ into the $d_x$ normalizing flows, $\{\G_{\btheta_d}(\cdot)\}_{d=1}^{d_x}$ and get the transformed GP function value $\f_t$, as shown in Eq.~\eqref{subeq:ETGPSSM_a}. Lastly, by conditioning on the sampled $\f_t$, we can sample the latent state $\x_t$ from $p(\x_t \vert \f_t)$.  In this way, we are able to numerically evaluate the first term, thereby enabling us to maximize the ELBO with respect to the variational parameters, $[\bzeta, \vz, \mathbf{m}, \mathbf{S}]$, and the model parameters $[\btheta_{gp}, \{\btheta_d\}_{d=1}^{d_x}, \mathbf{Q}, \mathbf{R}]$.
\begin{table}[t!]
    \begin{center}
        \caption{\footnotesize Comparisons between various variational GPSSMs. $c$ denotes the number of parameter shared by all models (including $[\bzeta, \mathbf{Q}, \mathbf{R}]$); $Q$ is the number of latent GPs in the ODGPSSM, which is typically $\ge d_x$; $\eta$ represents the parameter counts associated with the normalizing flows. For instance, in the case of a 2-layer SAL flow, $\eta=8$. \vspace{-.05in} }
        \label{tab:comparisons}
        \begin{adjustbox}{width=.49\textwidth}
        \setlength{\tabcolsep}{1.5mm}{
        \begin{tabular}{rclc}
            \toprule
             &\textbf{comput. complexity}   & \makecell[c]{\textbf{$\#$ parameters}}  &  \makecell[c]{\textbf{output-dependent}}  \\
            \midrule
            PRSSM \cite{doerr2018probabilistic}
            &  $\mathcal{O}(\cred{d_x} Tm^2)$
            &  $c + \cred{d_x} |\btheta_{gp}| +  {m \cred{d_x}(2 d_x \!+\! m \!+\! 4)}/{2}$ 
            &  \ding{55} \\ 
            ODGPSSM \cite{lin2022output}
            &  $\mathcal{O}(\cred{Q} Tm^2)$
            &  $c + \cred{Q} |\btheta_{gp}| + {m\cred{Q}(2 Q \!+\! m \!+\! 4)}/{2} + Qd_x$
            &  \ding{51}   \\ 
            EGPSSM (ours)
            &  $\mathcal{O}(Tm^2)$
            &  $c + |\btheta_{gp}| + {m(2d_x\!+\! m \!+\!4)}/{2} + \eta d_x $
            &  \ding{51} \\
            \bottomrule
        \end{tabular}}
        \end{adjustbox}
    \end{center}
    \vspace{-.35in}
\end{table}

So far, we have introduced the proposed method, encompassing efficient modeling and inference for GPSSM. We can compare the proposed method with some existing approaches, as summarized in Table \ref{tab:comparisons}. It is imperative to highlight that, in contrast to the existing approaches, our method significantly mitigates the escalating computational burden and parameter proliferation, resulting in enhanced efficiency for both modeling and inference. Meanwhile, the proposed method also establishes entangled relationships among the outputs because of the shared Gaussian copula \cite{wilson2010copula}.

\vspace{-.1in}
\section{Numerical Experiments}
\label{sec:experimental-results} \vspace{-.1in}

In this section, we present the performance evaluation of the proposed efficient GPSSM, referred to as EGPSSM, across synthetic and real datasets.  Given that the superiority of GPSSM over traditional time-series modeling methods has been shown in \cite{doerr2018probabilistic, lin2022output}, we will bypass redundant comparisons due to space limitations. Thus for the benchmark comparison, we consider two GPSSMs:  1)  PRSSM \cite{doerr2018probabilistic}; and 2)  ODGPSSM \cite{lin2022output}.

\vspace{-.12in}
\subsection{Synthetic Data} \label{subsec:synthetic_data}
\vspace{-.05in}
\begin{table}[t!]
    \begin{center}
        \caption{Prediction RMSE comparison between the proposed EGPSSM and the competitors on synthetic sequence.  Shown are mean and standard errors over five repetitions.}
        \label{tab:rmse-syn}
        \begin{adjustbox}{width=0.48\textwidth}
        \setlength{\tabcolsep}{3mm}{
        \begin{tabular}{cccc}
            \toprule
            \begin{tabular}[c]{@{}c@{}} PRSSM \end{tabular} & 
            \begin{tabular}[c]{@{}c@{}} ODGPSSM \end{tabular} &
            \begin{tabular}[c]{@{}c@{}} EGPSSM (L) \end{tabular} &
            \begin{tabular}[c]{@{}c@{}} EGPSSM (SAL) \end{tabular} \\
            \midrule
               $0.389 \!\pm\! 0.022$
            &  $0.388 \!\pm\! 0.031$
            &  $\bm{0.387 \!\pm\! 0.029}$
            &  $0.392 \!\pm\! 0.023$  \\
            \bottomrule
        \end{tabular}
        }
        \end{adjustbox}
    \end{center}
    \vspace{-.3in}
\end{table}
\begin{figure}[t!]
    \centering		
    \subfloat[$\#$ parameters vs. $d_x$]
    {
    \label{subfig:num_parameter_comparisons}
    \includegraphics[width =.465\linewidth]{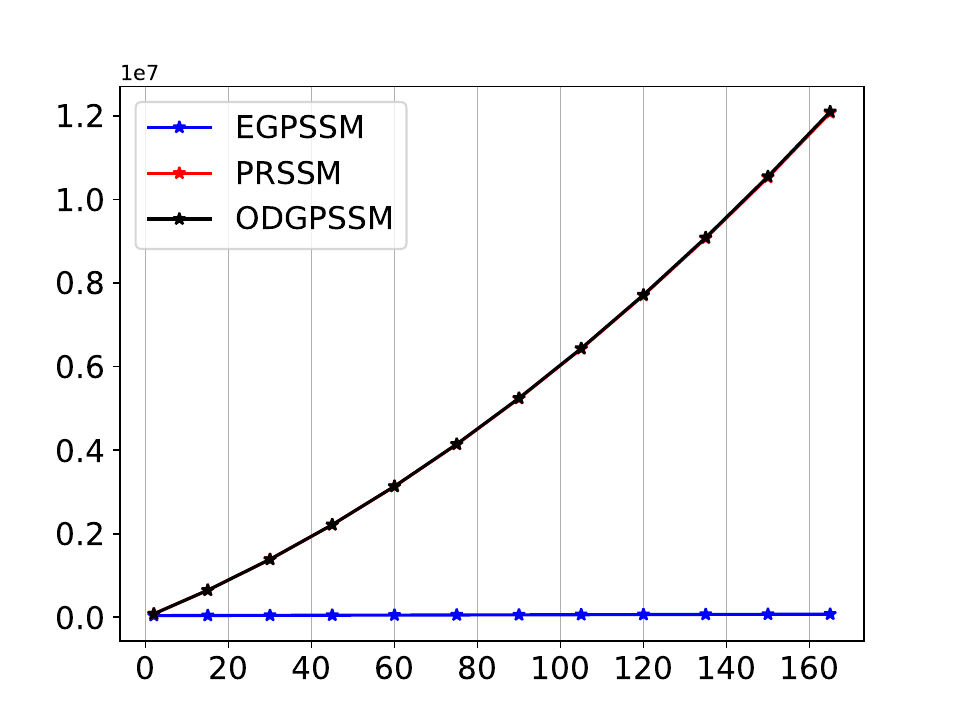}
    } \hfill 
    \subfloat[evaluation time (in seconds) vs. $d_x$]
    {
    \label{subfig:time_comparisons}
    \includegraphics[width =.475\linewidth]{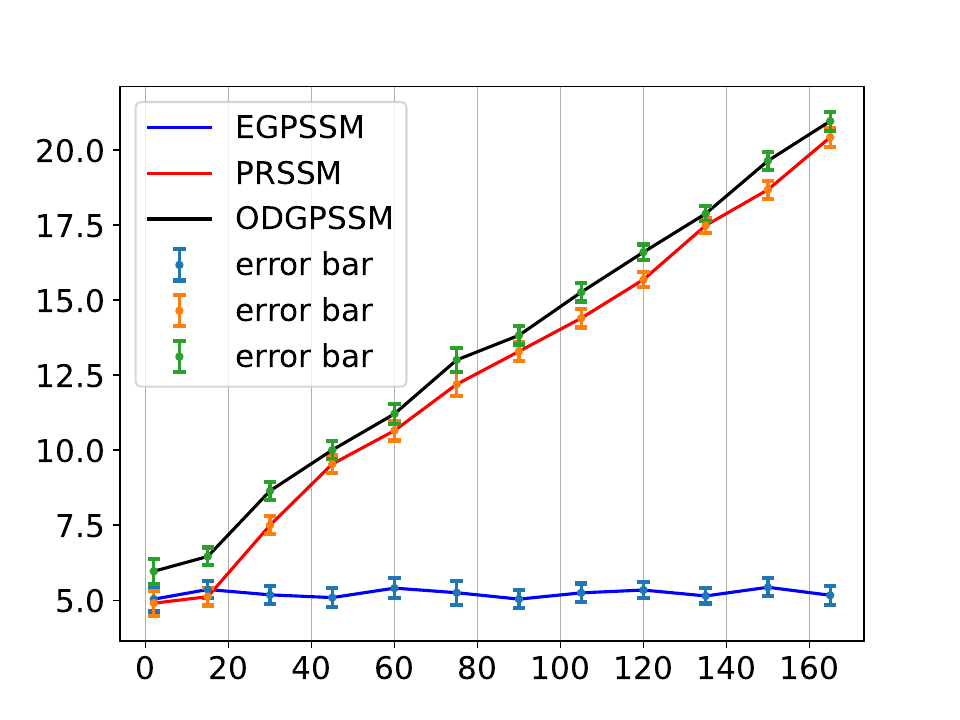}
    }
    \vspace{-.09in}
    \caption{Complexity comparisons between  different GPSSMs}
    \vspace{-.25in}
    \label{fig:EGPSSM_results}
\end{figure}
This subsection adopts the underlying two-dimensional SSM described below as the testing model,
\begin{subequations}
\label{eq:2dkink}
\setlength{\abovedisplayskip}{4pt}
\setlength{\belowdisplayskip}{5pt}
\begin{align}
    & f(\x_t) \!=\! 0.8 \!+\! \left(\x_{t,1}+0.2\right)\left[1 \!-\! \frac{5}{1+ \exp(-2\x_{t,1})}\right] + \x_{t, 2},\\
    & \x_{t+1} \!=\! 
    \begin{bmatrix}
        f(\x_t)\\
        -0.5 * f(\x_t)
    \end{bmatrix} \!+\! \x_{t} \!+\! \mathbf{v}_t,  \ \ \mathbf{v}_t \!\sim\! \cN(\bm{0}, \sqrt{0.001} \ \mathbf{I}_2), \label{subeq:2dkink}\\
    & \y_t \!=\! \x_t + \mathbf{e}_t,   \quad  \mathbf{e}_t \sim \cN(\bm{0}, \sqrt{0.01} \mathbf{I}_2),
\end{align}
\end{subequations}
where $f(\x_t)$ is a modified kink function that is commonly used in the GPSSM literature \cite{ialongo2019overcoming}.
We commence by showcasing the computational complexity and the number of parameters of different GPSSMs across different latent state dimensions. The corresponding results are illustrated in Fig.~\ref{fig:EGPSSM_results}. 
The computational time presented in Fig.~\ref{subfig:time_comparisons} is the wall time required for the GPSSMs to evaluate the ELBO once on a sequence with a length of 200. The results are obtained from five repeated experiments and displayed with mean and standard deviations. All the GPSSMs employ 200 inducing points and utilize the Mat\'{e}rn kernel \cite{williams2006gaussian}; EGPSSM integrates a 2-layer SAL flow. See our implemented code for more details.

Based on the results presented in Fig.~\ref{fig:EGPSSM_results}, it is evident that both PRSSM and ODGPSSM suffer from the escalating complexity. For example, PRSSM and ODGPSSM exhibit a substantial number of model parameters, nearly reaching 2 million, even when operating within a moderate latent state dimension of approximately 40. This significant parameter volume poses formidable optimization challenges. In contrast, EGPSSM offers distinct advantages over PRSSM and ODGPSSM, primarily attributed to its utilization of a shared GP and normalizing flow with a small number of parameters. This design choice contributes to enhanced computational efficiency and more efficient parameter management within the EGPSSM framework.

Subsequently, we set the latent state dimension to 2 for all GPSSMs, encompassing PRSSM, ODGPSSM, linear-flow EGPSSM (denoted as EGPSSM (L)), and SAL-flow EGPSSM (denoted as EGPSSM (SAL)).  These models are then deployed for sequence prediction tasks. In this context, we generated 10 training sequences of length 50, along with an additional test sequence of the same length. Our objective is to ascertain whether EGPSSM exhibits favorable inference performance.   The corresponding prediction root-mean-square error (RMSE) results are presented in Table~\ref{tab:rmse-syn}, revealing a notably comparable performance across all approaches. Notably, EGPSSM (L) displays a slightly superior performance, likely because it matches the linear relationship inherent in the transition function outputs of the underlying SSM, as shown in Eq.~\eqref{subeq:2dkink}. 
It is also worth noting that the existing GPSSMs possess sufficient model capability to accurately predict the sequence, although their intrinsic transition modeling is inconsistent with the underlying one and entails higher computational complexity and parameter count than EGPSSM.

\vspace{-.12in}
\subsection{Real Data} \label{subsec:read_data}
\vspace{-.08in}
\begin{table}[t!]
    \begin{center}
        \caption{Prediction RMSE comparison between the proposed EGPSSM and the competitors on the five system identification datasets. Shown are mean and standard errors over five repetitions. \vspace{-.1in}}
        \label{tab:rmse-SIdata}
        \begin{adjustbox}{width=0.5\textwidth}
        \setlength{\tabcolsep}{1mm}{
        \begin{tabular}{cccccc}
            \toprule
            Methods & 
            \begin{tabular}[c]{@{}c@{}} Actuator \end{tabular} & 
            \begin{tabular}[c]{@{}c@{}} Ballbeam \end{tabular} &
            \begin{tabular}[c]{@{}c@{}} Drive \end{tabular} &
            \begin{tabular}[c]{@{}c@{}} Dryer \end{tabular} & 
            \begin{tabular}[c]{@{}c@{}} Gas Furnace \end{tabular} \\
            \midrule
            PRSSM 
            &  $0.691 \!\pm\! 0.148$
            &  $0.074 \!\pm\! 0.010$
            &  $\bm{0.647 \!\pm\! 0.057}$
            &  $0.174 \!\pm\! 0.013$  
            &  $\bm{1.503 \!\pm\! 0.196}$   \\ 
            ODGPSSM 
            &  $\bm{0.666 \!\pm\! 0.074}$
            &  $0.068 \!\pm\! 0.006$
            &  $0.708 \!\pm\! 0.052$ 
            &  $\bm{0.171 \!\pm\! 0.011}$  
            &  $1.704 \!\pm\! 0.560$ \\
            EGPSSM (L)  
            &  $0.742 \!\pm\! 0.050$
            &  $\bm{0.055 \!\pm\! 0.005}$
            &  $0.756 \!\pm\! 0.020$
            &  $0.482 \!\pm\! 0.027$
            &  $1.994 \!\pm\! 0.085$ \\ 
            EGPSSM (SAL)  
            &  $0.758 \!\pm\! 0.048$
            &  $\bm{0.054 \!\pm\! 0.003}$
            &  $0.762 \!\pm\! 0.020$
            &  $0.479 \!\pm\! 0.009$
            &  $2.010 \!\pm\! 0.071$ \\ 
            \bottomrule
        \end{tabular}
        }
        \end{adjustbox}
    \end{center}
    \vspace{-.35in}
\end{table}
This subsection presents a comparative analysis involving EGPSSM, PRSSM, and ODGPSSM across five real system identification datasets, as detailed in \cite{doerr2018probabilistic}. 
In each dataset, the initial half of the sequence serves as training data, while the latter portion is allocated for testing purposes. Standardization of all datasets is performed using the training sequence, and a consistent latent state dimension of $d_x \!=\! 4$ is employed.
The SAL-flow EGPSSM incorporates a 2-layer SAL flow.  Further elaboration on the specific parameters is accessible through the associated online code repository. The prediction results are reported in Table \ref{tab:rmse-SIdata}, wherein the RMSE is averaged over a 50-step forecasting. Table~\ref{tab:rmse-SIdata} illustrates that, generally, EGPSSMs closely match the performance of the comparative models on the real datasets, with some instances even surpassing PRSSM and ODGPSSM, as seen in the \textit{Ballbeam} dataset. 
However, for certain datasets, EGPSSMs exhibit a slightly marginal disadvantage in sequence prediction.  This can predominantly be attributed to the incorporation of elementary flows in this paper, which introduces the potential challenge of limited model flexibility. Nonetheless, it is essential to acknowledge that EGPSSMs achieve comparable performance, while effectively managing model parameters and ensuring computational efficiency, particularly in settings characterized by moderate and high dimensional latent states.  
Subsequent research endeavors could focus on augmenting the model flexibility in the EGPSSM while preserving its efficiency. One prospective avenue for further improvement involves integrating more sophisticated normalizing flows to transform the shared GP. However,  delving into this matter, also, requires theoretical investigation into how the outputs of the ETGP correlate when using more complex flows. These aspects will be left for future investigations.

\vspace{-.13in}
\section{Conclusion} \label{sec:conclusion}
\vspace{-.13in}
This paper introduces a novel and efficient GPSSM, EGPSSM, tailored for modeling dynamical systems with high-dimensional latent states. Unlike existing approaches, our method can effectively mitigate the challenges of escalating computational complexity and parameter proliferation. Empirical evaluations, conducted across diverse synthetic and real-world datasets, substantiate the merits of EGPSSM, revealing its superior modeling efficiency and significant reduction in parameter count. Furthermore, our results demonstrate that EGPSSM achieves comparable performance to existing GPSSMs, underscoring its competitiveness in inference tasks. Nevertheless, the model flexibility might be somewhat constrained in intricate scenarios. As such, future work should focus on enhancing model flexibility while preserving computational efficiency.

\clearpage
\vfill\pagebreak

\bibliographystyle{IEEEtran}
\bibliography{refs-etgpssm}

\end{document}